\documentclass[letterpaper]{article} %
\usepackage{xai4sci}  %
\usepackage{times}  %
\usepackage{helvet}  %
\usepackage{courier}  %
\usepackage[hyphens]{url}  %
\usepackage{graphicx} %
\usepackage{natbib}  %
\usepackage{caption} %
\usepackage{algorithm}
\usepackage{algorithmic}
\usepackage{booktabs}

\usepackage{newfloat}
\usepackage{listings}
\DeclareCaptionStyle{ruled}{labelfont=normalfont,labelsep=colon,strut=off} %
\floatstyle{ruled}
\newfloat{listing}{tb}{lst}{}
\floatname{listing}{Listing}
\title{Data Science with LLMs and Interpretable Models}
\author {
    Sebastian Bordt\textsuperscript{\rm 1}\thanks{Work done at Microsoft Research.},
    Ben Lengerich\textsuperscript{\rm 2},
    Harsha Nori\textsuperscript{\rm 3},
    Rich Caruana\textsuperscript{\rm 3}
}
\affiliations {
    \textsuperscript{\rm 1}University of Tübingen, Tübingen AI Center\\
    \textsuperscript{\rm 2}Massachusetts Institute of Technology, Broad Institute of MIT and Harvard\\
    \textsuperscript{\rm 3}Microsoft Research\\
    sebastian.bordt@uni-tuebingen.de, blengeri@mit.edu, hanori@microsoft.com, rcaruana@microsoft.com
}

\usepackage{bibentry}

\begin{document}

\maketitle

\begin{abstract}
Recent years have seen important advances in the building of interpretable models, machine learning models that are designed to be easily understood by humans. In this work, we show that large language models (LLMs) are remarkably good at working with interpretable models, too. In particular, we show that LLMs can describe, interpret, and debug Generalized Additive Models (GAMs). Combining the flexibility of LLMs with the breadth of statistical patterns accurately described by GAMs enables dataset summarization, question answering, and model critique. LLMs can also improve the interaction between domain experts and interpretable models, and generate hypotheses about the underlying phenomenon. We release \url{https://github.com/interpretml/TalkToEBM} as an open-source LLM-GAM interface.   
\end{abstract}

\section{Introduction}

Recent years have seen important advances in the building of interpretable models \citep{caruana2015intelligible,lin2020generalized}. On tabular machine learning problems, interpretable glass-box models are now frequently on par with black box methods \citep{rudin2019stop}. In other words, today's interpretable models act as powerful pattern detectors. 

In this work, we show that combining interpretable models with large language models (LLMs) holds broad potential to improve many different tasks in data science. Using an interpretable model allows LLMs to handle complex statistical patterns, enabling them to perform some data science tasks semi-autonomously \cite{hollmann2023gpt}. For example, we can ask the LLM whether the statistical relationships learned by an interpretable model are surprising given what would be expected from prior knowledge in the underlying real-world problem.

We explore this approach with the interpretable model class of Generalized Additive Models (GAMs) \cite{hastie1990generalized,caruana2015intelligible}. GAMs represent complex outcomes as a sum of univariate component functions (graphs). This means that the model is additive, and we can make use of this additivity to scale our approach to large datasets with many features. By analyzing the graphs of a GAM one at a time, the LLM can analyze large models with many different features, without ever having the entire model fit into the context window \cite{liu2023lost}. This is the  advantage of using interpretable models: they have short textual description lengths. In contrast, black-box models are monolithic and have long textual descriptions, prohibiting their use as contexts for LLMs.\footnote{The textual description length of the GAM on the pneumonia dataset %
is 43,592 GPT-4 tokens, too many to be directly encoded in GPT-4's 32k context window. The individual graphs are much more compact: the maximum description length of a single graph in the pneumonia GAM is 2,345 GPT-4 tokens. This easily fits within GPT-4's context window.}

In addition to serving as a useful tool for data science, the combination of LLMs with interpretable models also holds the potential to improve domain expert's interaction with machine learning. This is because it provides domain experts with textual access to the models, and allows them to formulate questions to the model in natural language \cite{slack2022talktomodel}. %

This paper is organized as follows. We first provide a brief overview of our approach. We then provide simple baseline evaluations (is the LLM able to read a value from a graph?), and continue with examples of more complex qualitative tasks (summarization, anomaly detection). We then discusses the problem of hallucination, and presents some simple experimental evidence on the grounding of model responses in the interpretable model. %

\section{Our Approach}
\label{sec:approach}

\begin{figure*}
    \centering
    \includegraphics[width=\textwidth]{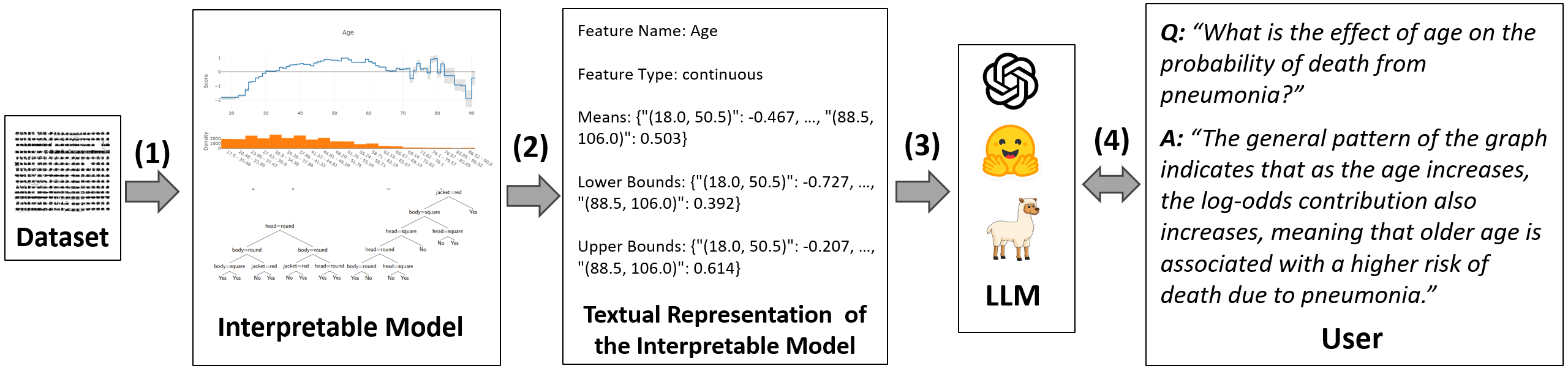}
    \caption{Conceptual overview of our approach. (1) We first train an interpretable model. (2) Next, we convert the different components of the interpretable model to text. (3) Then we provide these components as inputs to an LLM. (4) The user  asks the LLM questions about the model or the dataset. The LLM can provide model-level summaries, or answer specific questions about particular aspects of the model. The LLM can also be used to automatically generate hypotheses about the real-world phenomenon that underlies the model and data.}
    \label{fig:conceptual overview}
\end{figure*}

The short technical description of our approach is as follows. We first train a GAM. Then we convert the individual graphs of the GAM to text, representing them as key-value pairs in \texttt{JSON}.
The graphs are then provided as context to an LLM, together with a general introductory prompt about the task, and a description of the dataset. We  engage in chain-of-thought (CoT) prompting \cite{wei2022chain}, first asking the LLM to describe the general pattern of the graph, then asking more specific questions (these questions depend on the task that we want the LLM to perform). We also request succinct summarizations of entire conversations about graphs. To elevate the discussion to the level of the entire model and dataset, the summaries of the conversations about individual graphs are aggregated and provided as context in a new query to the LLM. A graphical overview of this approach is given in Figure \ref{fig:conceptual overview}. The task of describing an individual graph is depicted in Figure \ref{fig:describe graph}. The prompt structure is given in the Supplement.

\section{Experimental Results}
\label{sec:results}

This sections provides the results of different experiments. We begin by investigating the ability of GPT-3.5 and GPT-4 to perform baseline tasks with graphs. Performance on these tasks can be quantified. This leads us to the conclusion that our approach requires relatively powerful LLMs. We then demonstrate the impressive abilities of GPT-4 to perform complex qualitative tasks with GAMs.

{\bf Datasets.} We use the California Housing \cite{KELLEYPACE1997291}, OpenML Diabetes \cite{smith1988using,vanschoren2014openml}, Iris, Kaggle Titanic\footnote{\url{https://www.kaggle.com/competitions/titanic}}, Kaggle Spaceship Titanic\footnote{\url{https://www.kaggle.com/c/spaceship-titanic}}, and Pneumonia \cite{caruana2015intelligible} dataset. The Spaceship Titanic dataset is from 2022, and the Pneumonia dataset has never been released on the Internet, meaning that GPT-3.5 and GPT-4 have not seen these datasets during training (the pneumonia dataset has been described in papers).

{\bf Interpretable Models.} We train Generalized Additive Models (GAMs) using the \texttt{interpretml} software package \cite{nori2019interpretml}.

{\bf LLMs.} We perform our experiments with \texttt{gpt-3.5-turbo-16k-0613} and \texttt{gpt-4-0613}, using the OpenAI API \cite{openai2023gpt4}.

\subsection{Baseline Tasks}
\label{subsec:baselineTasks}

We conduct the following experiments using 31 different graphs, obtained from training GAMs on the different datasets. 

\begin{figure*}
    \centering
    \includegraphics[width=\textwidth]{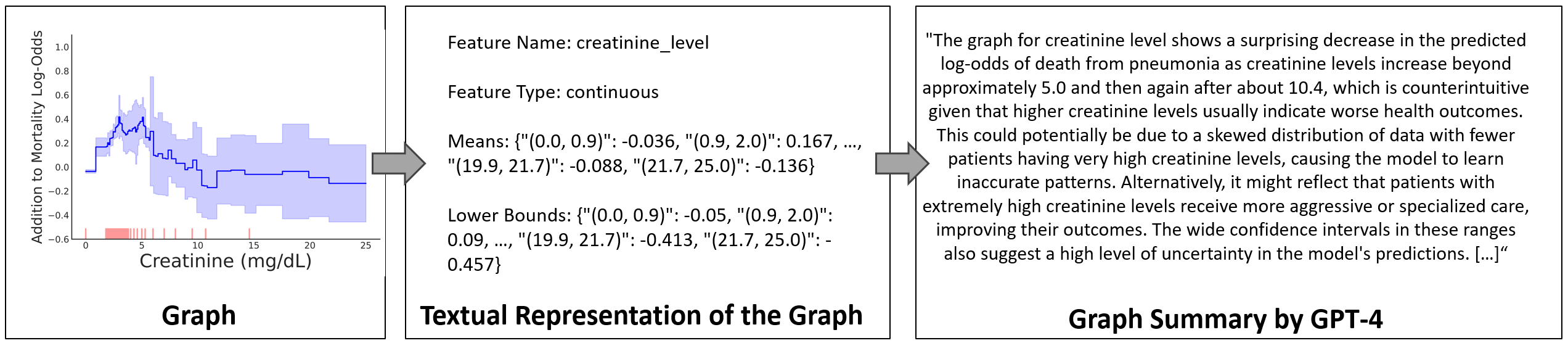}
    \caption{The basic building block of our framework is the ability of LLMs to describe and summarize the individual graphs of  Generalized Additive Models (GAMs). Considering a GAM one graph at a time allows the LLM to work with interpretable models even on large-scale datasets with hundreds of features while staying within confined context windows.}
    \label{fig:describe graph}
\end{figure*}

{\bf Reading a Value from a Graph.} As a test of whether the LLM is able to perform basic tasks with the \texttt{JSON} format in which we provide the graph, we ask the LLM to read a value from the graph ("What is the mean value of the graph at 3.528?"). This means that we sample a random point on the x-axis and ask the LLM about the value of the graph at that point. The results of this experiment are depicted in Table \ref{tab:baseline-table}. GPT-4 correctly reads 64 out of 75 values from the different graphs. In fact, the success rate of GPT-4 on some graphs is much higher, and the mistakes frequently occur on graphs that require many decimal places to encode. GPT-3.5, in contrast, is not able to perform this simple task (of course, this can also depend on prompting).

 {\bf Deciding Monotonicity.} One of the most important properties that a graph in an interpretable model can have is monotonicity. If a graph is monotone, it would be important to include this in a summarization of the graph, and later the summary of the full model. We test the ability of GPT-3.5 and GPT-4 to decide whether a graph is monotone ("Please tell me whether the graph is monotone increasing, monotone decreasing, or not monotone."). By using both the original and y-axis inverted graphs, we conduct this experiment with a total of 62 different graphs (if the original graph is monotone increasing, then the y-axis inverted graph is monotone decreasing).
The result of this experiment is depicted in the second row of Table \ref{tab:baseline-table}. GPT-4 correctly identifies 11 of 12 monotone graphs. It also claims that 9 out of 50 non-monotone graphs are monotone. However, most of these graphs 9 are ``honest mistakes'', meaning that they are almost, but not perfectly, monotone. We thus find that GPT-4 has a surprisingly strong ability to decide the monotonicity of a graph purely from its textual representation. GPT-3.5, in contrast, is hardly able to identify the monotone graphs.

{\bf Finding the Largest Jump.} Inspired by the strong performance of GPT-4 on the previous two tasks, we decided to ask the model to identify the largest jump in a graph. The graphs of EBMs are piecewise constant, which means that they have jumps in between the different intervals \cite{nori2019interpretml}. The learned jumps often have real-world interpretations \cite{lengerich2022death}. However, this task is challenging for the LLMs since a perfect solution requires taking the max absolute value of the subtraction of adjacent values \cite{dziri2023faith}. The result of the experiment is depicted in the third column of Table \ref{tab:baseline-table}. We ask the LLMs to identify the biggest jumps in our 31 different graphs. Using chain-of-thought prompting (first listing a number of important jumps in the graph, then listing the largest one) GPT-4 is able to achieve an impressive success rate of 27 correctly identified jumps out of a total of 31 graphs. Again, the number of  jumps identified by GPT-3.5 is much smaller.

\subsection{Qualitative Tasks}

In the previous Section, we saw that GPT-4 shows good performance on simple tasks with graphs. In this Section, we provide different examples that demonstrate the impressive capabilities of GPT-4 on more complex qualitative tasks.

{\bf Summarizing Graphs.} A basic building block of our framework is the ability of GPT-4 to describe and summarize individual graphs. Here we ask the model to describe the influence of the categorical feature ``Home Planet'' in the fictional Spaceship Titanic task from Kaggle. The model predicts the probability of a passenger ``being transported to an alternative dimension''.

\begin{quote}
    \textbf{GPT-4:}\it "The HomePlanet feature has a significant influence on the probability of a passenger being transported to an alternate dimension during the collision with the spacetime anomaly. Passengers from Europa have the highest positive effect, indicating a higher likelihood of being transported. On the other hand, passengers from Earth have a negative effect, suggesting a lower likelihood of being transported. [...]"
\end{quote}

{\bf Summarizing Models.} Based on the ability of LLMs to describe individual graphs, we can use them to describe entire models. For this task, we additionally provide the LLM with the global feature importances.
Consider the following description of a GAM trained to predict the likelihood of death from pneumonia \cite{caruana2015intelligible}.

\begin{quote}
    \textbf{GPT-4:}\it "The model predicts the likelihood of death from pneumonia based on several patient features. The most influential global feature is age, with increasing age associated with a higher likelihood of death, particularly after the age of 73. Respiration rate and BUN level also have substantial influences on the model's predictions, with higher levels of both associated with an increased risk of death.

Other significant features include the number of diseases a patient has, their systolic blood pressure, and their pO2 levels. More diseases, lower systolic blood pressure, and very low or very high pO2 levels are all associated with a higher likelihood of death.

The model also shows that certain medical readings can increase or decrease the likelihood of death, depending on their levels. For example, both very low and very high glucose levels, pH levels, and potassium levels can increase the risk of death, while moderate levels are associated with a lower risk.

Finally, the model shows a considerable amount of uncertainty in its predictions for certain features, such as age, heart rate, and albumin levels. This is likely due to a lack of data points in certain ranges or the complex, non-linear relationships these features may have with the outcome. [...]"
\end{quote}

This description was generated using the approach outlined above.
Any aspects mentioned in the description are entirely correct and grounded in the actual model. (We discuss the problem of hallucination in the next Section).

\begin{table}
  \caption{Performance of GPT-3.5 and  GPT-4 on baseline tasks with graphs. The table depicts the number of cases where the model succeeded versus the total number of cases in the task.}
  \label{tab:baseline-table}
  \centering
  \begin{tabular}{lll}
    \toprule
     & GPT-3.5 & GPT-4 \\
    \midrule
    Reading a Value from a Graph & 3/75  & 64/75     \\
    Deciding Monotonicity     &  2/12$\quad$ & 11/12$\quad$      \\
    Finding the Largest Jump     & 9/31       & 27/31  \\
    \bottomrule
  \end{tabular}
\end{table}

{\bf Anomaly Detection.} In order to show how the combination of LLMs and interpretable models can be useful for practitioners, we demonstrate the ability of GPT-4 at anomaly detection. In this task, we ask the LLM to critique the model and find any aspects that might be surprising or anti-causal \cite{lengerich2022death}. We again use the pneumonia model and ask the LLM to find surprising patterns in the graph for the feature ``creatinine level''.

\begin{quote}
    \textbf{GPT-4:} \it The graph for creatinine level shows a surprising decrease in the predicted log-odds of death from pneumonia as creatinine levels increase beyond approximately 5.0 and then again after about 10.4, which is counterintuitive given that higher creatinine levels usually indicate worse health outcomes. This could potentially be due to a
skewed distribution of data with fewer patients having very high creatinine levels, causing the model to learn inaccurate patterns. Alternatively, it might reflect that patients with extremely high creatinine levels receive more aggressive or specialized care, improving their outcomes. The wide confidence intervals in these ranges also suggest a high level of uncertainty in the model's predictions. [...]"
\end{quote}

A clinician rated this response as  accurate and potentially useful. For a more detailed analysis of how our approach can be useful for anomaly and surprise detection in healthcare, see \cite{lengerich2023llms}.

\section{Grounding and Hallucination}
\label{sec:hallucination}

In this Section, we conduct an empirical investigation of the problem of hallucination. Our approach is to modify the learned graphs in a way that is counterfactual to that found in the original data.
We then ask the LLM to describe both the original and counterfactual graph. In the absence of hallucination, the description of the counterfactual graph should reflect the modifications. We first discuss results with individual graphs. We then discuss the task of describing a counterfactual model.

\subsection{Describing a Counterfactual Graph}

We first discuss the example of the ``Sex'' feature on the Kaggle Titanic dataset. The reason why we chose this example because the LLM has likely learned about the role of sex in emergency evacuations during pre-training, which might lead it to ignore the graph and provide a generic response. Here is GPT-4 describing the graph of the ``Sex'' feature in a GAM that models the survival probability on the Titanic.

\begin{quote}
\textbf{GPT-4:} \it "The 'Sex' feature significantly influenced the survival outcome in the Titanic disaster. Being female greatly increased the chances of survival, as indicated by a positive mean value of 1.397. [..] the overall trend suggests that females had a higher survival rate compared to males, aligning with the historical context of the "women and children first" maritime evacuation policy."
\end{quote}

We can see how the LLM provides an interpretation within the historical context. Now, we flip the values of 'male' and 'female' in the model, and see how the description of the LLM changes.

\begin{quote}
\textbf{GPT-4:} \it "Being male contributes positively to the log-odds of survival, with a mean contribution of 1.397, while being female contributes negatively [...] it should be noted that these do not represent raw survival probabilities. This counterintuitive finding may be influenced by other confounding variables in the dataset. Further investigation is needed to fully understand this output."
\end{quote}

We observe that the model response changes in accordance with the graph. Interestingly, the LLM also notes that the result is counterintuitive, even if it does not directly explain why this is the case (which would be even better and might be achieved with better prompting).

\subsection{Describing a Counterfactual Model}

We now discuss an experiment where GPT-4 describes a counterfactual model. We use the popular Iris dataset. The reason why we chose this dataset is because GPT-4 has seen it during pre-training \citep{bordt2023elephants}, meaning that GPT-4's responses could potentially be based on the original data and not on our interpretable model. We first ask GPT-4 to describe a GAM that models the probability of a flower being Iris-setosa.

\begin{quote}
\textbf{GPT-4:} \it "[...] Iris-setosa flowers typically have shorter and narrower petals compared to other species. [...] As sepal length increases beyond 5.55 units, the likelihood of a flower being Iris-setosa decreases. However, Iris-setosa flowers are more likely to have wider sepals, with the likelihood increasing drastically for widths greater than 2.85 units. Each feature exhibits an abrupt change in contribution at specific thresholds, suggesting potential distinct cutoffs that differentiate Iris-setosa from other species. [...]
\end{quote}

We then invert the graphs of the GAM on the y-axis and again pretend to the LLM that this GAM predicts the probability of a flower being Iris-setosa. Here is the description of the modified model.

\begin{quote}
\textbf{GPT-4:} \it    "[...] The model suggests that Iris Setosa is more likely to have longer sepals, smaller sepal$\ $widths, longer petal lengths, and wider petals. Notably, there are threshold effects at certain intervals for all the features. [...] Further validation with expert biological knowledge or additional data would be beneficial."
\end{quote}

We observe that the model summarization by the LLM changes in accordance with the model.

In summary, while the evidence from our simple qualitative experiments is far from conclusive, they do demonstrate that the responses of GPT-4 are frequently grounded in the provided interpretable model.

\section{Related Work}
\label{sec:related_work}
 
Slack et al. \cite{slack2022talktomodel} develop a natural language interface to give practitioners conversational access to model explanations. Their work is similar in spirit to ours but intends to provide access to generic (potentially black-box) models, precluding the LLM from direct access to model internals. Bisercic et al. \cite{bisercic2023interpretable} showed that LLMs can extract tabular datasets from unstructured text and then train interpretable models (linear regression and small decision trees) on top of this data. Various recent works have explored the potential of LLMs for data science with tabular data by data wrangling and cleaning \cite{narayan2022can,vos2022towards}, or traditional supervised and unsupervised learning tasks like classification or density estimation \citep{borisov2022language,hegselmann2023tabllm,wang2023anypredict,onishi2023tabret}. %

Previous work with GAMs has shown the ubiquity of hidden confounding in real-world datasets and the utility of GAMs to reveal surprising effects in datasets \cite{caruana2015intelligible,lengerich2022death}. 
This prior work relied on human experts to design explicit statistical tests for surprising patterns in GAM feature graphs, limiting the potential applications.

\section{Discussion}
\label{sec:discussion}

In this work we have explored synergies between LLMs and interpretable models and found this to be a powerful combination with broad potential to assist data scientists and practitioners.

As a preliminary investigation, we explored the capabilities of GPT-3.5 and GPT-4 to perform a number of different tasks with graphs generated by glass-box models. Although we gave only limited effort to prompt tuning, the results already are reasonably strong and probably could be further improved with more advanced prompting strategies. 
We observe that it would be unclear how to give an LLM the same degree of access to a black-box model. 
In principle, however, a related approach might be possible with post-hoc methods, for example, Shapley Values, which are known to be closely related to GAMs \citep{lundberg2017unified,bordt2023shapley}. %

Our preliminary experiments raise a variety of avenues of future study. 
For example, how does LLM  performance change with graph complexity? Under what conditions is the LLM response grounded in the graph, and when does hallucination occur?

It is important to note that, given the current state of the technology, there are no guarantees that an answer from an LLM is a truthful response to the prompt. 
For this reason, all the answers and descriptions provided by the LLMs should be seen as helpful suggestions rather than final answers. Similarly, the answers provided by GPT-4 in the anomaly detection tasks should be seen as useful hypothesis that have to be verified by human experts.

All in all, we believe that the approach of combining LLMs with interpretable models holds broad potential to assist practitioners and domain experts across a broad range of tasks.

\bibliography{surprise}

\onecolumn
\appendix
\section{Appendix}

\begin{figure*}[h!]
    \centering
    \fbox{\begin{minipage}{0.9\textwidth}\footnotesize
\input{prompts/describe_graph_1}
    \end{minipage}}
    \label{fig:apx prompt structure}
    \caption{(continues on the next page)}
\end{figure*}

\newpage
\begin{figure*}[h!]
    \centering
    \fbox{\begin{minipage}{0.9\textwidth}\footnotesize
\input{prompts/describe_graph_2}
    \end{minipage}}
    \caption{The prompt structure to describe graphs that is the basis for the results in this paper. In the depicted example, we ask the model to describe the Age graph of a GAM trained on the Titanic dataset from Kaggle. Note that we provide the LLM with a description of the dataset and the meaning of the values on the y-axis in the graphs. The prompt structure is fairly general and can be easily amended to better fit various specific use cases. }
\end{figure*}

\end{document}